# Piecewise Linear Units Improve Deep Neural Networks

Jordan Inturrisi, Sui Yang Khoo, Abbas Kouzani
School of Engineering, Deakin University, Australia
{jinturri, sui.khoo, abbas.kouzani}@deakin.edu.au

Riccardo Pagliarella
riccardo.pagliarella@gmail.com

## Abstract

*The activation function is at the heart of a deep neural networks nonlinearity; the choice of the function has great impact on the success of training. Currently, many practitioners prefer the Rectified Linear Unit (ReLU) due to its simplicity and reliability, despite its few drawbacks. While most previous functions proposed to supplant ReLU have been hand-designed, recent work on learning the function during training has shown promising results. In this paper we propose an adaptive piecewise linear activation function, the Piecewise Linear Unit (PiLU), which can be learned independently for each dimension of the neural network. We demonstrate how PiLU is a generalised rectifier unit and note its similarities with the Adaptive Piecewise Linear Units, namely adaptive and piecewise linear. Across a distribution of 30 experiments, we show that for the same model architecture, hyperparameters, and pre-processing, PiLU significantly outperforms ReLU: reducing classification error by 18.53% on CIFAR-10 and 13.13% on CIFAR-100, for a minor increase in the number of neurons. Further work should be dedicated to exploring generalised piecewise linear units, as well as verifying these results across other challenging domains and larger problems.*

## 1. Introduction

Machine learning, particularly the emergence of deep learning, has been gaining momentum as techniques capable of consuming vast amounts of data in order to construct highly complex nonlinear abstractions [1]. The goal of machine learning is to iteratively learn relationships hidden within data without being explicitly programmed [1]. Deep learning specifically stacks multiple layers to produce a highly nonlinear output.

At the heart of every deep neural network is a linear transformation followed by a nonlinear activation function. Typically, the weights of the linear components are learned through an algorithm such as gradient descent, while the nonlinearity is predetermined when architecting the model. While an infinitely large neural network is believed to be able to approximate arbitrarily complex functions [2], the choice of nonlinearity has a major impact on the success of training a network of finite size.

Currently, practitioners prefer the Rectified Linear Unit (ReLU), defined as $f(x) = \max(0, x)$ [3, 4], as it is mathematically simple and computationally efficient – deep rectifier networks achieved superior performance on supervised tasks without requiring unsupervised pre-training [3, 5-8]. When $x > 0$, ReLU ensures no vanishing/exploding gradient through an identity derivative – effectively propagating error gradients through the active paths of neurons, leading to faster learning and better convergence compared to sigmoid or tanh units.

Nevertheless, for ReLU to learn symmetric and antisymmetric underlying characteristics of data, it would require twice as many neurons as symmetric/antisymmetric activation functions [7]. Since ReLU is non-negative, it has a non-zero mean activation which causes a bias shift for the next layer – the greater the magnitude of the mean activation, the higher the bias shift [9]. This negatively affects performance. Moreover, when $x \leq 0$, $f'(x) = 0$, leading to cases where a neuron may never activate since gradient-based optimisation will never make a weight update [10]. On the other hand, during training a weight update may cause the unit to never activate on any data point again. Both these cases are colloquially known as *"dead neurons"*. Like the result of the vanishing gradient issue, when training ReLU with constant 0 gradients, training would plateau, leading to poorer convergence [10].

Over the years, numerous activation functions have been proposed to improve upon ReLU – most notably Leaky ReLU [9-11], Parametric ReLU [11], Exponential Linear Units (ELU) [9], Swish [12], Adaptive Piecewise Linear (APL) units [13], PLU [14] and others [15-18]. Nevertheless, none have managed to gain the same level of favour as ReLU; the performance improvements of other functions are inconsistent across different models and datasets. In saying that, most papers are limited in their comparisons: either comparing results after 1 run, or the median of 5 runs. Understandably these results would be noisy and likely the cause for inconsistent performances. Rather, it would be prudent to compare a performance distribution, say across 30 runs, when comparing performance results.

Typically, these competing activation functions are hand-designed according to desirable properties, though the space of possible functions is vast. Recently, search techniques have been effective in other domains in discovering traditionally hand-designed components [12, 13, 19-21]. Taking this approach, we could explore the space of functions by *learning* the activation function during training.

In this work we propose two adaptive piecewise linear activation functions which can be learned independently for each network dimension, using gradient descent – DoubleReLU and the Piecewise Linear Unit (PiLU). Both proposed functions are related to but differ from APL units [13] and PLUs [14]. We focus on scalar piecewise linear activation functions since they can be substituted for ReLU without altering the network, while their output is linear with the input – avoiding any chaotic outputs from nonlinear functions. Across a distribution of 30 experiments, we show that for the same model architecture, hyperparameters, and pre-processing, PiLU significantly outperforms ReLU: reducing classification error by 18.53% on CIFAR-10 and 13.13% on CIFAR-100, for a minor increase in the number of neurons.

## 2. Piecewise Linear Functions

### 2.1. DoubleReLU

DoubleReLU is similar to threshold linear functions applied to unsupervised autoencoders in [22, 23], and also draws similarities to hard thresholding [24]. DoubleReLU creates a 'zone' of true zeroes between $x = \pm\alpha$; though the output is not thresholded and starts from zero outside this zone. More formally, DoubleReLU is defined as

$$f(x) = \begin{cases} x - \alpha & \text{when } x > \alpha \\ 0 & \text{when } -\alpha \leq x \leq \alpha \\ x + \alpha & \text{when } x < -\alpha \end{cases}$$

Resulting in distinctly separate derivatives

$$f'(x) = \begin{cases} 1 & \text{when } |x| > |\alpha| \\ 0 & \text{elsewhere} \end{cases}$$

Therefore, providing a linear derivative with a zone of true zeroes and an inherent nonlinearity. In this zone of true zeroes, the derivative is zero.

DoubleReLU is like standard ReLU, though extended to allow negative values in the output, hence the mean activation can be centred around 0. Figure 1 plots the graph of DoubleReLU for different values of $\alpha$. If $\alpha = 0$, DoubleReLU becomes the identity function; as $\alpha \to \infty$, the zone of true zeroes increases in size. DoubleReLU can be viewed as a linear function with a zone which requires some magnitude of pre-activation to activate.

Unlike ReLU and other rectifiers, DoubleReLU is unbounded both above and below the origin.

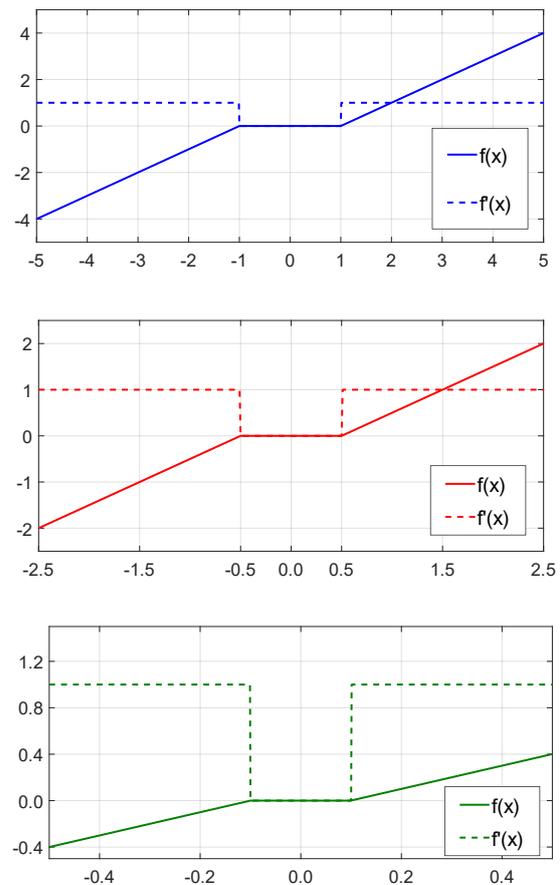

*Figure 1: DoubleReLU function and first derivative: top) α=1.0, middle) α=0.5, bottom) α=0.1*

### 2.2. Piecewise Linear Unit (PiLU)

Taking inspiration from current state-of-the-art ReLU-based activation functions, we hypothesise the benefit is achieved via the (quasi) linear piecewise nature of these functions [13]. These functions include ReLU, PReLU, ELU, RReLU, APL, PLU etc. We propose to further generalise PReLU by providing an adaptive knot along with 2 adaptive gradients either side of the knot.

We want a function to be unbounded and piecewise linear to ensure outputs don't explode due to high nonlinearity (outputs are linear with inputs), yet the function overall is nonlinear as in typical rectifier functions. We call this function the Piecewise Linear Unit (PiLU); it is given by

$$f(x) = \begin{cases} \alpha x + \gamma(1-\alpha) & \text{when } x > \gamma \\ \beta x + \gamma(1-\beta) & \text{when } x \leq \gamma \end{cases}$$

Resulting in 2 distinctly separate derivatives

$$f'(x) = \begin{cases} \alpha & \text{when } x > \gamma \\ \beta & \text{when } x \leq \gamma \end{cases}$$

The result is a generalisation of PReLU which can learn a piecewise linear function with a knot occurring at γ. This introduces 3 additional parameters per function compared with standard ReLU; an additional 2 parameters compared with standard PReLU. The benefit here is a linear output based on the input, with a nonlinearity introduced via the knot. Nonetheless, there is no guarantee this function will result with a zero mean activation or be symmetric about the origin.

### 2.3. PiLU: The generalised rectifier unit

It is trivial to show that PiLU is a generalisation of ReLU and hence by extension, LReLU, PReLU, and other rectifier-based units; ReLU and other rectifier-based units are a special-case of PiLU.

For ReLU, LReLU, and PReLU:

- $\gamma = 0$, i.e. the "knot" is at zero; and,
- $\alpha = 1$, i.e. the output is linear above the knot; and,
- ReLU: $\beta = 0$, LReLU: $\beta = 0.01$, and PReLU: $\beta = \delta$ (where δ is the adaptive parameter).

Using the parameters in Table 1, we can produce ReLU, LReLU, and PReLU from the PiLU definition.

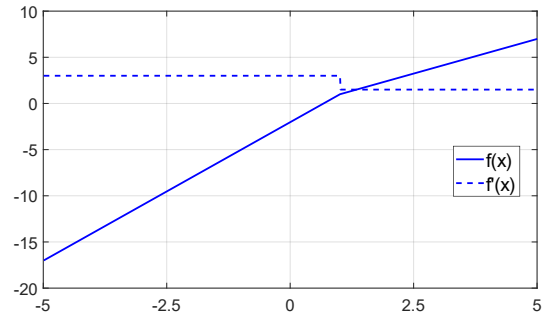

*Figure 2: PiLU with α=1.5, β=3, and γ=1*

### 2.4. PiLU: Complexity

It is worth commenting on the additional complexity of the novel activation functions. Clearly the space complexity of ReLU is $O(0)$ as there are no adaptive weights; PReLU, DoubleReLU, and PiLU are all linear with $O(n)$ though total space depends on the total number of adaptive weights.

The time complexity for ReLU, PReLU, DoubleReLU, and PiLU are linear with $O(n)$ (Figure 3). Though, it is obvious that the more complex functions – in order of complexity: ReLU → PReLU → DoubleReLU → PiLU – require additional computation time (Figure 4). As noted, this additional computation scales linearly with the input size (tested with input vector of sizes: 100, 1,000, 10,000, and 100,000 across 100,000 iterations). As the input increases, the relative differences in computation time decreases compared to ReLU (Figure 4).

*Table 1: PiLU comparison to rectifier units*

|  | **ReLU** | **LReLU** | **PReLU** |
|---|---|---|---|
| Parameters | $\alpha = 1, \beta = 0, \gamma = 0$ | $\alpha = 1, \beta = 0.1, \gamma = 0$ | $\alpha = 1, \beta = \delta, \gamma = 0$ |
| Simplified Equation | $f(x) = \begin{cases} x & \text{when } x > 0 \\ 0 & \text{when } x \leq 0 \end{cases}$ | $f(x) = \begin{cases} x & \text{when } x > 0 \\ 0.01x & \text{when } x \leq 0 \end{cases}$ | $f(x) = \begin{cases} x & \text{when } x > 0 \\ \delta x & \text{when } x \leq 0 \end{cases}$ |
| Derivatives | $f'(x) = \begin{cases} 1 & \text{when } x > 0 \\ 0 & \text{when } x \leq 0 \end{cases}$ | $f'(x) = \begin{cases} 1 & \text{when } x > 0 \\ 0.01 & \text{when } x \leq 0 \end{cases}$ | $f'(x) = \begin{cases} 1 & \text{when } x > 0 \\ \delta & \text{when } x \leq 0 \end{cases}$ |

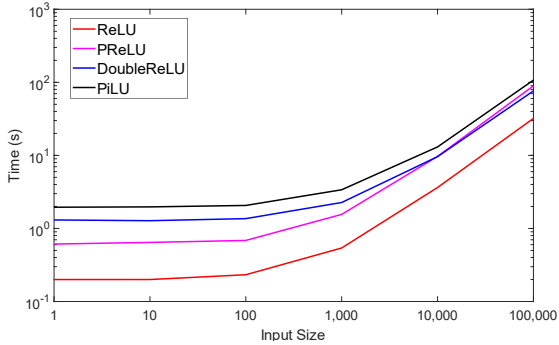

Figure 3: Activation Function Time Complexity

## 3. Experimental Setup

To test the robustness of activation functions, we run each test across 30 different random seeds.

### 3.1. Model Architecture

Much time can be spent optimising the model architecture in search of state-of-the-art results. Indeed, many researchers and papers are dedicated to this area of research. In this investigation, apart from using some well-developed intuitions in computer vision architectures [25, 26], little time is spent in optimising the model architecture. This is so we can focus our efforts on comparing the difference in performance based on the difference in activation function. Concretely, we used the model architecture in Table 2 for both the CIFAR-10 and CIFAR-100 datasets.

For the following experiments, we apply a channel-wise adaptive weight sharing scheme – for activation functions with adaptive weights – whereby the adaptive weight is shared across per channel for each layer. Thus, the number of additional parameters per layer is equal to the product of the number of adaptive weights (denoted $n$ in Table 2) and the number of channels. The increase in parameters is negligible when considering the total number of parameters. We anticipate no risk in overfitting.

### 3.2. Hyperparameters

The Glorot normal weight initialisation scheme [27] was used at each layer for initialisation. L2 regularisation of weight 1e-3 was applied to the fully connected output layer. The loss function used was the categorical cross entropy (negative log likelihood). The Adam optimisation algorithm was used with $\alpha = 0.001$, $\beta_1 = 0.9$, $\beta_2 = 0.999$, and $\varepsilon = 10^{-7}$ [28].

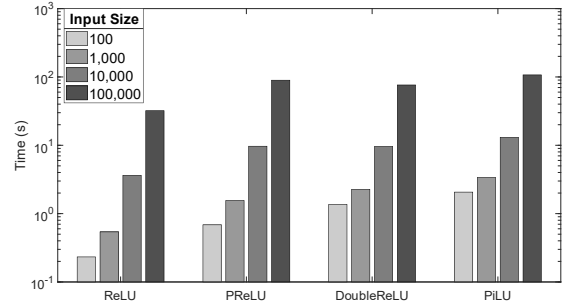

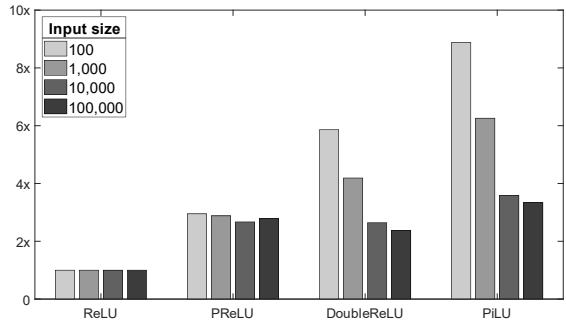

Figure 4: Comparison of computation time for various input sizes: top) Total computation time, bottom) Computation time compared to ReLU

We extend the approach of [27]; throughout training all metrics, weights, gradients, and outputs are monitored and logged after each epoch – for each of the training, validation, and test set. This incurs considerable computation and storage costs (as there are a lot of numbers!), though is useful for any ensuing analysis. Each simulation was run across 30 random seeds, so we could develop a distribution of results. We use the same hyperparameters for each simulation. All programming was completed in Python, making use of the packages NumPy [29], Keras [30], and Tensorflow [31]. We applied the same model architecture as in Table 2, though replacing the activation function as needed.

### 3.3. Data Pre-Processing

The input values are normalised to between 0 and 1. Since 8-bit RGB values are represented by integer values between 0 and 255, normalisation between 0 and 1 can be achieved by dividing the value by 255. The output values are converted from dense class vectors to sparse class matrices. For both CIFAR-10 and CIFAR-100, the validation set is taken as the last 10,000 images of the training set. This leaves 40,000 images as part of the training set, 10,000 images as part of the validation set, and 10,000 images as part of the test set. Each dataset has approximately uniform sampling from each of the 10/100 classes.

# 4. Results

Since the only variance between models for each random seed value is the change in activation function, any statistical differences in results can be attributed to such function differences. This allows an analysis of different activation functions on performance, convergence, and training time. Baseline models trained with linear, ReLU, and PReLU neurons are used for comparison. Since there are 30 experiments for each model (each experiment with a different seed value), we can observe the performance of the models from a range of initialisations. From this, we develop a distribution of values to better indicate the range of performance. Statistical comparisons will be used to quantify model performance and compare between functions.

Now we compare the performance of the novel activation functions (DoubleReLU and PiLU) against the typical state-of-the-art (i.e., ReLU and PReLU) functions. To validate the effectiveness of the novel activation functions, we empirically evaluate them against CIFAR-10 and CIFAR-100 datasets [32]. Under the channel-wise scheme, the adaptive weights are set for each (convolutional) channel in each layer. Hence, few additional parameters are added (Table 3).

## 4.1. CIFAR-10

After training for 50 epochs, the loss is minimised, and accuracy maximised, on the novel activation function PiLU, followed in second place by PReLU, then DoubleReLU and ReLU last (Figure 5). PiLU improves on PReLU and ReLU accuracy by 1.93 and 6.2 percentage points, respectively. This improved performance is from a 0.89% and 1.34% increase in parameters compared to PReLU and ReLU, respectively.

Figure 5 shows the distribution of activation function test set metrics (both loss and accuracy) on CIFAR-10, composed of results from 30 random seeds. It shows, regardless of the adaptive weight scheme, PiLU is the best performing activation function across 50 epochs, followed by PReLU. DoubleReLU outperforms ReLU for the Channel- and Neuron-wise adaptive weight schemes, whilst their performance is approximately equal for the Layer-wise scheme.

## 4.2. CIFAR-100

After training for 50 epochs, the loss is minimised, and accuracy maximised, on the novel activation function PiLU, followed in equal second-place of PReLU and DoubleReLU, with ReLU last (

Figure 6). PiLU improves on PReLU and ReLU accuracy by 3.05 and 9.55 percentage points, respectively. This improved performance is from a 0.77% and 1.15% increase in parameters compared to PReLU and ReLU, respectively.

*Table 2: Model Architecture*

| Input size | Output size | Layer | Parameters |
|---|---|---|---|
| 32x32x3 | 30x30x16 | 3x3, 16 CONV2D (Layer 1) | 448 |
| 30x30x16 | 30x30x16 | Activation Function | $16n$ |
| 30x30x16 | 15x15x16 | 2x2, Max Pool | 0 |
| 15x15x16 | 13x13x16 | 3x3, 16 CONV2D (Layer 2) | 2320 |
| 13x13x16 | 13x13x16 | Activation Function | $16n$ |
| 13x13x16 | 11x11x32 | 3x3, 32 CONV2D (Layer 3) | 4640 |
| 11x11x32 | 11x11x32 | Activation Function | $32n$ |
| 11x11x32 | 9x9x32 | 3x3, 32 CONV2D (Layer 4) | 9248 |
| 9x9x32 | 9x9x32 | Activation Function | $32n$ |
| 9x9x32 | 7x7x64 | 3x3, 64 CONV2D (Layer 5) | 18496 |
| 7x7x64 | 7x7x64 | Activation Function | $64n$ |
| 7x7x64 | 64 | Average Pooling 2D | 0 |
| 64 | 64 | Dropout(p=0.5) | 0 |
| 64 | {10/100} | {10/100}, Fully Connected | {650/6500} |
| {10/100} | {10/100} | SoftMax Activation Function | 0 |

*Table 3: Comparison of activation function parameters with ReLU*

|  | CIFAR-10 | | | CIFAR-100 | | |
|---|---|---|---|---|---|---|
|  | **Parameters** | **# Increase** | **% Increase** | **Parameters** | **# Increase** | **% Increase** |
| ReLU | 35,802 | - | - | 41,652 | - | - |
| PReLU | 35,962 | 160 | 0.45% | 41,812 | 160 | 0.38% |
| DoubleReLU | 35,962 | 160 | 0.45% | 41,812 | 160 | 0.38% |
| PiLU | 36,282 | 480 | 1.34% | 42,132 | 480 | 1.15% |

Figure 6 shows similar results for the distribution of activation function test set metrics on CIFAR-100. Again, regardless of the adaptive weight scheme, PiLU is the best performing activation function across 50 epochs. Though on the more difficult CIFAR-100, the results of DoubleReLU and PReLU are approximately equal under the Channel-wise adaptive weight scheme, with PReLU slightly outperforming on the Layer- and Neuron-wise schemes. Under all schemes, ReLU is the worst performing activation function.

## 5. Discussions and Conclusions

We proposed two new piecewise linear activation functions – DoubleReLU and PiLU – with desirable properties to overcome the vanishing gradient problem. Their adaptive parameters are learned via gradient descent alongside the typical weight parameters. We applied these novel activations functions against CIFAR-10 and CIFAR-100 and compared to current state-of-the-art ReLU and PReLU to investigate their effectiveness. Each experiment was repeated across 30 random seeds to produce a statistical distribution of performance, providing further clarity of performance differences. Our experiments demonstrate that learning the function during training can improve classification performance with a negligible increase in number of parameters.

Across both CIFAR-10 and CIFAR-100 datasets, we showed the novel PiLU outperforms current state-of-the-art ReLU and PReLU with minimal increases in the number of parameters. Our extensive experiments show that PiLU consistently outperforms ReLU and PReLU, at least in image classification domains. On CIFAR-10, replacing ReLUs with PiLUs improves classification error by 18.53% (6.2 percentage points) for a slight increase in parameters of 1.34% (480 parameters); on CIFAR-100, replacing ReLUs with PiLUs improves classification error by 13.13% (9.55 percentage points) for a slight increase in parameters of 1.15% (480 parameters). These improvements are both significant and statistically significant.

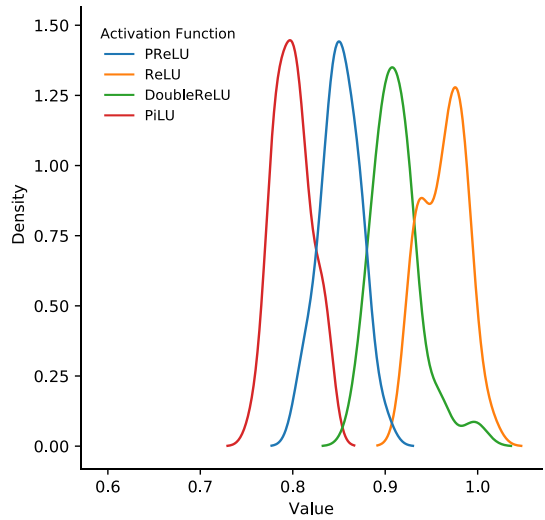

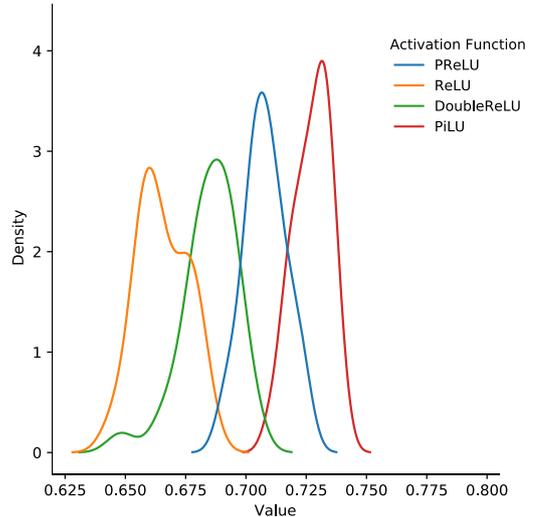

*Figure 5: Distribution of Activation Function metrics (CIFAR-10 – Channel-wise): top) Loss, bottom) Acuracy*

*Table 4: Activation function test set performance on CIFAR-10*

|            | Accuracy         | Error            | Loss              | Parameters |
|------------|------------------|------------------|-------------------|------------|
| ReLU       | 66.54 ± 0.38%    | 33.46 ± 0.38%    | 0.964 ± 0.0082    | 35,802     |
| PReLU      | 70.81 ± 0.31%    | 29.19 ± 0.31%    | 0.851 ± 0.0075    | 35,962     |
| DoubleReLU | 68.49 ± 0.41%    | 31.51 ± 0.41%    | 0.912 ± 0.0092    | 35,962     |
| PiLU       | 72.74 ± 0.27%    | 27.26 ± 0.27%    | 0.799 ± 0.0071    | 36,282     |

*Table 5: Activation function test set performance on CIFAR-100*

|            | Accuracy         | Error            | Loss              | Parameters |
|------------|------------------|------------------|-------------------|------------|
| ReLU       | 27.26 ± 0.35%    | 72.74 ± 0.38%    | 2.987 ± 0.0176    | 41,652     |
| PReLU      | 33.76 ± 0.23%    | 66.24 ± 0.31%    | 2.646± 0.0086     | 41,812     |
| DoubleReLU | 33.39 ± 0.31%    | 66.61 ± 0.41%    | 2.664 ± 0.0129    | 41,812     |
| PiLU       | 36.81 ± 0.17%    | 63.19 ± 0.27%    | 2.491 ± 0.0079    | 42,132     |

Ultimately, PiLU achieves a better local minimum in the loss landscape faster than ReLU and PReLU. This revelation is important, particularly since a change in function, with minimal additional parameters, leads to a vast improvement in performance. Our results suggest that the standard approach of manually prespecifying an activation function may be suboptimal.

## 6. Future Work

Much of the prior innovations in deep neural network activation function have focused on proposing new activation functions [9-11, 13, 17, 33, 34]. Few studies such as [35], [12] and this one have systematically compared different activation functions. Further, since the typical approach is to report either the best result or the median of 5 results, we believe this is one of very few studies to compare a distribution of the learning metric's performance for scalar activation functions.

This paper provides the groundwork, and suggests some directions, for future work in neural network activation functions:

1. **Generalised piecewise linear units:** Future investigations should further extend this view of adaptive piecewise linear activation functions through multiple knots, similar to Adaptive Piecewise Linear (APL) units [13] and PLUs [14]. The current trajectory of activation function innovation has been leading in this direction, and given by the results presented throughout this paper, this should continue. There are no doubt further techniques to improve upon the PiLU function we have proposed here. As mentioned, search techniques have been shown to be effective at automating the discovery of traditionally hand-designed components [12, 19-21].

2. **More than just scalar-to-scalar activation functions:** while this paper has focused on scalar activation functions, there are other types used in deep neural networks that are worth exploring, including: *Many-to-one* (i.e. max-pooling, maxout [36], gating [37-42]), *One-to-many* (i.e. Concatenated ReLU [43]) and *Many-to-many* (i.e. BatchNorm [44], LayerNorm [45]) functions.

3. **Other challenging domains, larger problems:** while the results presented in this paper are promising, it is unclear (though likely) that PiLU can successfully replace ReLU on datasets in other domains and real-world datasets (e.g. machine translation, speech recognition, etc.), as well as within much larger networks (e.g. networks required for ImageNet).

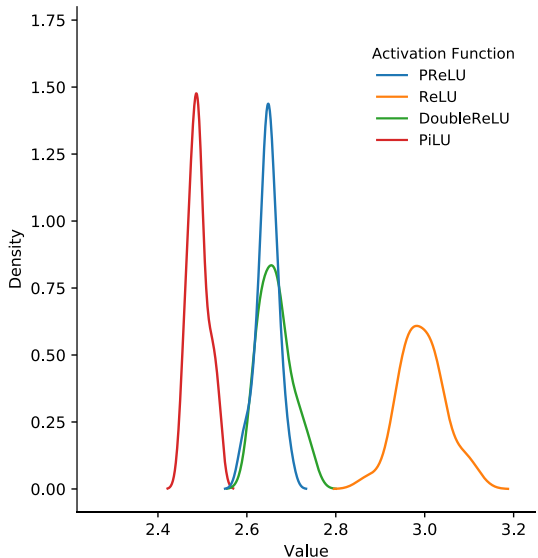

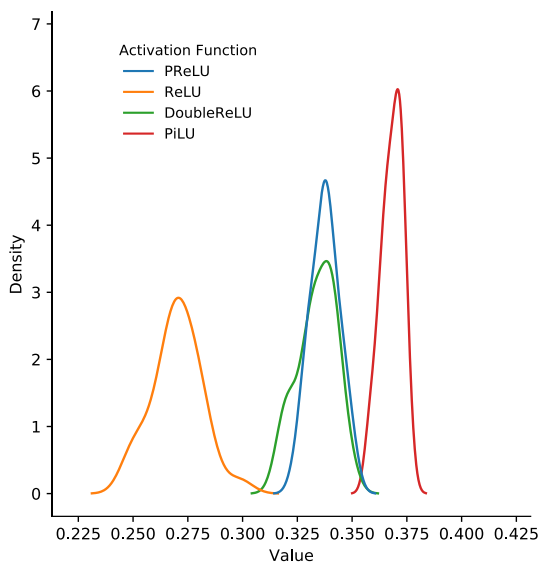

*Figure 6: Distribution of Activation Function Metrics (CIFAR-100 – Channel-wise): top) Loss, bottom) Accuracy*

## Acknowledgements

We thank Deakin University for the postgraduate research scholarship which supported this research.

## References

[1] K. P. Murphy, *Machine Learning: A Probabilistic Perspective*. The MIT Press, 2012, p. 1096.

[2] K. Hornik, M. Stinchcombe, and H. White, "Multilayer feedforward networks are universal approximators," *Neural networks*, vol. 2, no. 5, pp. 359-366, 1989.

[3] Y. LeCun, Y. Bengio, and G. Hinton, "Deep learning," *Nature*, vol. 521, no. 7553, pp. 436-444, 2015.

[4] V. Nair and G. E. Hinton, "Rectified linear units improve restricted boltzmann machines," in *Proceedings of the 27th international conference on machine learning (ICML-10)*, 2010, pp. 807-814.

[5] A. Krizhevsky, I. Sutskever, and G. E. Hinton, "Imagenet classification with deep convolutional neural networks," in *Advances in neural information processing systems*, 2012, pp. 1097-1105.

[6] J. Schmidhuber, "Deep learning in neural networks: An overview," *Neural networks*, vol. 61, pp. 85-117, 2015.

[7] X. Glorot, A. Bordes, and Y. Bengio, "Deep sparse rectifier neural networks," in *Proceedings of the Fourteenth International Conference on Artificial Intelligence and Statistics*, 2011, pp. 315-323.

[8] G. E. Dahl, T. N. Sainath, and G. E. Hinton, "Improving deep neural networks for LVCSR using rectified linear units and dropout," in *2013 IEEE international conference on acoustics, speech and signal processing*, 2013, pp. 8609-8613: IEEE.

[9] D.-A. Clevert, T. Unterthiner, and S. Hochreiter, "Fast and accurate deep network learning by exponential linear units (elus)," *arXiv preprint arXiv:1511.07289,* 2015.

[10] A. L. Maas, A. Y. Hannun, and A. Y. Ng, "Rectifier nonlinearities improve neural network acoustic models," in *Proc. ICML*, 2013, vol. 30, no. 1.

[11] K. He, X. Zhang, S. Ren, and J. Sun, "Delving deep into rectifiers: Surpassing human-level performance on imagenet classification," in *Proceedings of the IEEE international conference on computer vision*, 2015, pp. 1026-1034.

[12] P. Ramachandran, B. Zoph, and Q. V. Le, "Searching for activation functions," *arXiv preprint arXiv:1710.05941,* 2017.

[13] F. Agostinelli, M. Hoffman, P. Sadowski, and P. Baldi, "Learning activation functions to improve deep neural networks," *arXiv preprint arXiv:1412.6830,* 2014.

[14] A. Nicolae, "PLU: The piecewise linear unit activation function," *arXiv preprint arXiv:1809.09534,* 2018.

[15] S. Kong and M. Takatsuka, "Hexpo: A vanishing-proof activation function," in *Neural Networks (IJCNN), 2017 International Joint Conference on*, 2017, pp. 2562-2567: IEEE.

[16] A. N. S. Njikam and H. Zhao, "A novel activation function for multilayer feed-forward neural networks," *Applied Intelligence,* vol. 45, no. 1, pp. 75-82, 2016.

[17] S. Qiu and B. Cai, "Flexible Rectified Linear Units for Improving Convolutional Neural Networks," *arXiv preprint arXiv:1706.08098,* 2017.

[18] L. Trottier, P. Giguère, and B. Chaib-draa, "Parametric exponential linear unit for deep convolutional neural networks," *arXiv preprint arXiv:1605.09332,* 2016.

[19] B. Zoph and Q. V. Le, "Neural architecture search with reinforcement learning," *arXiv preprint arXiv:1611.01578,* 2016.

[20] I. Bello, B. Zoph, V. Vasudevan, and Q. V. Le, "Neural optimizer search with reinforcement learning," in *International Conference on Machine Learning*, 2017, pp. 459-468: PMLR.

[21] B. Zoph, V. Vasudevan, J. Shlens, and Q. V. Le, "Learning transferable architectures for scalable image recognition," in *Proceedings of the IEEE conference on computer vision and pattern recognition*, 2018, pp. 8697-8710.

[22] R. Goroshin and Y. LeCun, "Saturating auto-encoders," *arXiv preprint arXiv:1301.3577,* 2013.

[23] R. Memisevic and D. Krueger, "Zero-bias autoencoders and the benefits of co-adapting features," *stat,* vol. 1050, p. 13, 2014.

[24] C. J. Rozell, D. H. Johnson, R. G. Baraniuk, and B. A. Olshausen, "Sparse coding via thresholding and local competition in neural circuits," *Neural computation,* vol. 20, no. 10, pp. 2526-2563, 2008.

[25] F. N. Iandola, S. Han, M. W. Moskewicz, K. Ashraf, W. J. Dally, and K. Keutzer, "SqueezeNet: AlexNet-level accuracy with 50x fewer parameters and< 0.5 MB model size," *arXiv preprint arXiv:1602.07360,* 2016.

[26] D. Mishkina, N. Sergievskiyb, and J. Matasa, "Systematic evaluation of CNN advances on the ImageNet," *Center for Machine Perception, Faculty of Electrical Engineering,* 2016.

[27] X. Glorot and Y. Bengio, "Understanding the difficulty of training deep feedforward neural networks," in *Proceedings of the Thirteenth International Conference on Artificial Intelligence and Statistics*, 2010, pp. 249-256.

[28] D. P. Kingma and J. Ba, "Adam: A method for stochastic optimization," *arXiv preprint arXiv:1412.6980,* 2014.

[29] C. R. Harris *et al.*, "Array programming with NumPy," *Nature,* vol. 585, no. 7825, pp. 357-362, 2020.

[30] F. Chollet. (2015). *Keras*. Available: https://keras.io

[31] M. Abadi *et al.*, "Tensorflow: A system for large-scale machine learning," in *12th {USENIX} symposium on operating systems design and implementation ({OSDI} 16)*, 2016, pp. 265-283.

[32] A. Krizhevsky and G. Hinton, "Learning multiple layers of features from tiny images," 2009.

[33] D. Hendrycks and K. Gimpel, "Bridging nonlinearities and stochastic regularizers with gaussian error linear units," 2016.

[34] S. Elfwing, E. Uchibe, and K. Doya, "Sigmoid-weighted linear units for neural network function approximation in reinforcement learning," *Neural Networks,* vol. 107, pp. 3-11, 2018.

[35] B. Xu, N. Wang, T. Chen, and M. Li, "Empirical evaluation of rectified activations in convolutional network," *arXiv preprint arXiv:1505.00853,* 2015.

[36] I. J. Goodfellow, D. Warde-Farley, M. Mirza, A. Courville, and Y. Bengio, "Maxout networks," *arXiv preprint arXiv:1302.4389,* 2013.

[37] A. v. d. Oord, N. Kalchbrenner, O. Vinyals, L. Espeholt, A. Graves, and K. Kavukcuoglu, "Conditional image generation with pixelcnn decoders," *arXiv preprint arXiv:1606.05328,* 2016.

[38] Y. N. Dauphin, A. Fan, M. Auli, and D. Grangier, "Language modeling with gated convolutional networks," in *International conference on machine learning*, 2017, pp. 933-941: PMLR.

[39] S. Hochreiter and J. Schmidhuber, "Long short-term memory," *Neural computation,* vol. 9, no. 8, pp. 1735-1780, 1997.

[40] R. K. Srivastava, K. Greff, and J. Schmidhuber, "Highway networks," *arXiv preprint arXiv:1505.00387,* 2015.

[41] J. Chung, C. Gulcehre, K. Cho, and Y. Bengio, "Empirical evaluation of gated recurrent neural networks on sequence modeling," *arXiv preprint arXiv:1412.3555,* 2014.

[42] J. Chung, C. Gülçehre, K. Cho, and Y. Bengio, "Gated Feedback Recurrent Neural Networks," in *ICML*, 2015, pp. 2067-2075.

[43] W. Shang, K. Sohn, D. Almeida, and H. Lee, "Understanding and improving convolutional neural networks via concatenated rectified linear units," in *international conference on machine learning*, 2016, pp. 2217-2225: PMLR.

[44] S. Ioffe and C. Szegedy, "Batch normalization: Accelerating deep network training by reducing internal covariate shift," in *International conference on machine learning*, 2015, pp. 448-456: PMLR.

[45] J. L. Ba, J. R. Kiros, and G. E. Hinton, "Layer normalization," *arXiv preprint arXiv:1607.06450,* 2016.